\documentclass[10pt,twocolumn,letterpaper]{article}
\usepackage{amsmath}
\usepackage[accsupp]{axessibility}
\usepackage[T1]{fontenc}
\usepackage{times}
\usepackage{cvpr}

\definecolor{cvprblue}{rgb}{0.21,0.49,0.74}
\usepackage[pagebackref,breaklinks,colorlinks,allcolors=cvprblue]{hyperref}
\usepackage[utf8]{inputenc}
\usepackage{booktabs}
\usepackage{multirow}
\usepackage{makecell}
\usepackage{siunitx}
\usepackage[table]{xcolor}
\definecolor{myorange}{RGB}{255,100,3}
\definecolor{mygray}{gray}{.85}
\definecolor{mygray1}{gray}{.7}
\definecolor{mygray2}{gray}{.93}
\definecolor{mygray3}{gray}{.90}
\usepackage{array}

\def\paperID{}
\def\confName{CVPR}
\def\confYear{2026}
\title{SymphoMotion: Joint Control of Camera Motion and Object Dynamics for Coherent Video Generation}

\author{
Guiyu Zhang\textsuperscript{1},  
Yabo Chen\textsuperscript{2},  
Xunzhi Xiang\textsuperscript{3}, \\
Junchao Huang\textsuperscript{1},  
Zhongyu Wang\textsuperscript{4},
Li Jiang\textsuperscript{\textdagger 1}\\
\textsuperscript{1}The Chinese University of Hong Kong, Shenzhen
\textsuperscript{2}Shanghai Jiaotong University \\
\textsuperscript{3}Nanjing University
\textsuperscript{4}Beihang University \\
{\tt\small guiyuzhang@link.cuhk.edu.cn, chenyabo@sjtu.edu.cn, xbxsxp@gmail.com,}\\
{\tt\small junchaohuang@link.cuhk.edu.cn, wangzhongyu@buaa.edu.cn, jiangli@cuhk.edu.cn.} \\
\url{https://grenoble-zhang.github.io/SymphoMotion/}
}

\begin{document}
\vspace{-10mm}
\twocolumn[{
\renewcommand\twocolumn[1][]{#1}
\maketitle
\footnotetext{\textsuperscript{*}Equal Contribution. \textsuperscript{\textdagger}Corresponding Author.}

\begin{center}
    \captionsetup{type=figure}
    \includegraphics[width=1\textwidth]{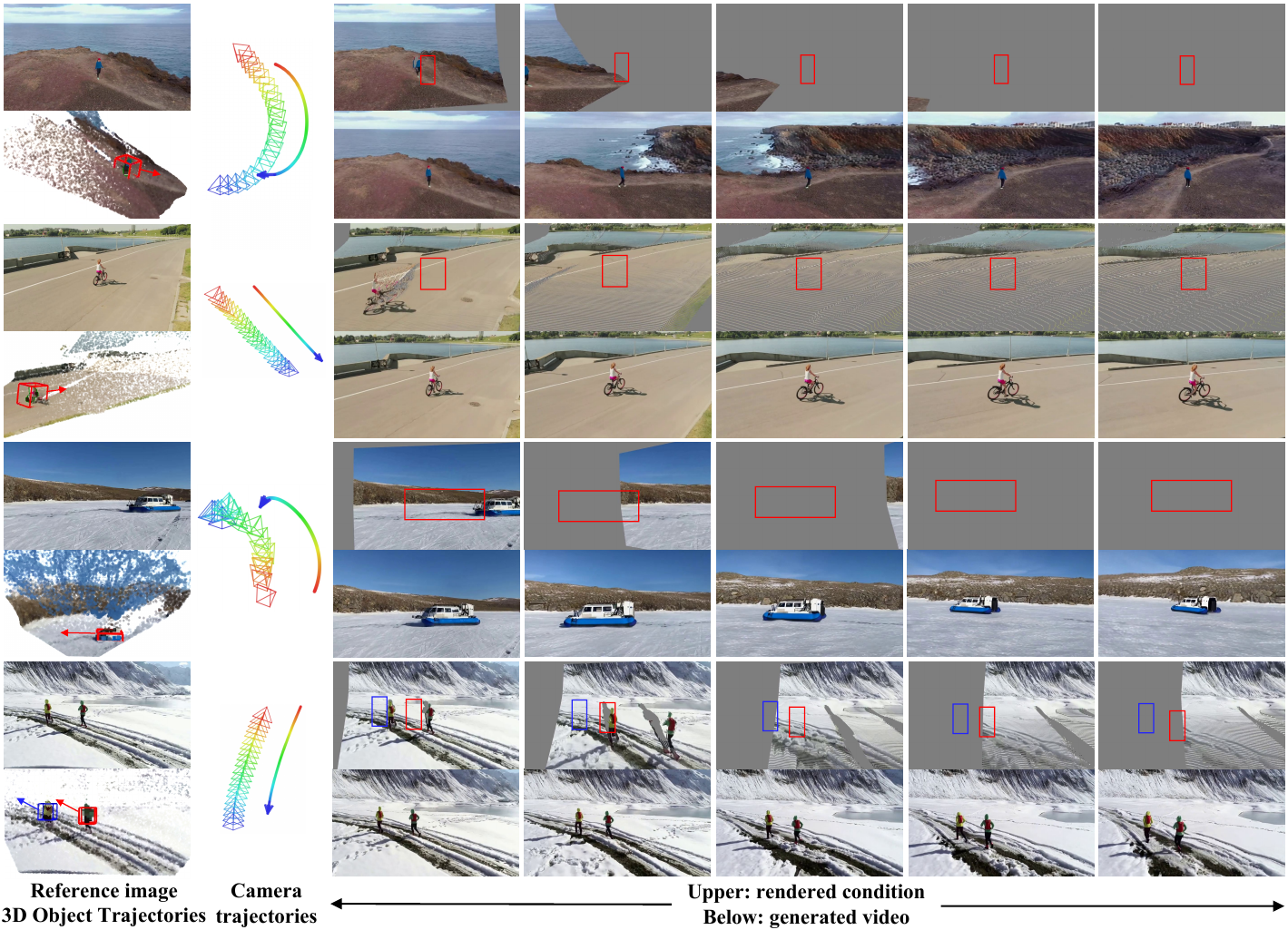}
    \vspace{-3mm}
    \caption{\textbf{Joint Control of Camera and Object Motion.} Given a reference image, a set of 3D object trajectories, and a camera trajectory, SymphoMotion generates videos that are spatially consistent and faithfully reflect both object and camera motion.}
    \label{fig:Teaser}
\end{center}
}]

\begin{abstract}
Controlling both camera motion and object dynamics is essential for coherent and expressive video generation, yet current methods typically handle only one motion type or rely on ambiguous 2D cues that entangle camera-induced parallax with true object movement. We present SymphoMotion, a unified motion-control framework that jointly governs camera trajectories and object dynamics within a single model. SymphoMotion features a Camera Trajectory Control mechanism that integrates explicit camera paths with geometry-aware cues to ensure stable, structurally consistent viewpoint transitions, and an Object Dynamics Control mechanism that combines 2D visual guidance with 3D trajectory embeddings to enable depth-aware, spatially coherent object manipulation. To support large-scale training and evaluation, we further construct RealCOD-25K, a comprehensive real-world dataset containing paired camera poses and object-level 3D trajectories across diverse indoor and outdoor scenes, addressing a key data gap in unified motion control. Extensive experiments and user studies show that SymphoMotion significantly outperforms existing methods in visual fidelity, camera controllability, and object-motion accuracy, establishing a new benchmark for unified motion control in video generation.

\end{abstract}

\section{Introduction}
\label{sec:intro}

{\noindent\itshape
“A symphony is like the world, it must contain everything.”%
\vspace{-0.5em}
\begin{flushright}
\normalfont--- Gustav Mahler
\end{flushright}
}
\vspace{-0.5em}

\noindent Precise control of motion dynamics in video generation has gained increasing attention~\cite{zhang2025proteus, hu2024motionmaster, chu2025wan, ling2024motionclone, huang2026live}, as it enables customized synthesis and richer visual expression. In filmmaking, directors coordinate camera movement and actor trajectories to shape narrative intent; analogously, controllable video generation requires jointly steering both camera motion and object dynamics to produce coherent and meaningful scenes. However, achieving such unified control remains challenging: camera trajectories induce global parallax and viewpoint changes, while objects follow independent, often complex 3D paths. Existing methods typically handle only one motion type, resulting in unsynchronized behaviors and reduced realism in naturally dynamic scenes.

Camera-control methods~\cite{he2024cameractrl,feng2024i2vcontrol,xu2024camco,hou2024training,bahmani2025ac3d,bahmani2024vd3d} generally inject camera parameters or view-related cues to regulate viewpoint transitions. While effective in static or near-static settings, these approaches model camera motion in isolation and are unable to capture how camera trajectories interact with moving objects, often degrading when significant foreground dynamics are present. Conversely, object-control methods~\cite{shi2024motion,yin2023dragnuwa,zhang2025tora,zhou2025trackgo,jain2024peekaboo,li2025magicmotion,fu20243dtrajmaster} rely primarily on 2D motion cues such as bounding boxes, trajectories, or optical flow. Such image-plane representations are inherently viewpoint-dependent and fail to disentangle true object motion from camera-induced parallax, making them unreliable under camera movement or large viewpoint changes.

Recent attempts toward joint control encode both camera and object motion within shared 2D motion fields or dense correspondences, such as the optical flow or point-trajectory representations employed by MotionPrompting~\cite{geng2025motion} and ATI~\cite{wang2025ati}. However, mixing camera-induced parallax and true object dynamics in the same 2D space leads to ambiguous supervision, as distinct 3D motions can project similarly onto the image plane, especially in scenes with substantial depth variation. Methods such as MotionCtrl~\cite{wang2024motionctrl} and Perception-as-Control~\cite{chen2025perception} move toward disentanglement by introducing separate processing branches for camera and object motion, yet they still define object trajectories purely in 2D image space, limiting their ability to model depth-aware motion or maintain consistency under strong camera movement. To address the inherent limitations of image-plane trajectory modeling, FMC~\cite{shuai2025free} uses explicit 6-DoF pose trajectories to represent motion in true 3D space; however, its reliance on synthetic data and the requirement for fully specified 6-DoF inputs hinder its practicality in real-world scenarios, where such detailed annotations are rarely available. These limitations underscore the need for a unified, 3D-aware, and intuitive representation capable of reliably guiding both camera and object motion.

To address these limitations, we propose SymphoMotion, a unified motion-control framework that jointly handles camera trajectories and dynamic object manipulation. SymphoMotion comprises two complementary mechanisms. The Camera Trajectory Control (CTC) enhances viewpoint control by combining explicit camera trajectories with geometry-aware cues that help preserve scene structure and maintain consistency throughout the generated sequence. The Object Dynamics Control (ODC) governs object dynamics by integrating 2D visual guidance with 3D trajectory embeddings, enabling objects to move along user-specified paths in full 3D space while remaining spatially coherent with the evolving viewpoint. In addition, SymphoMotion provides flexible interfaces for specifying motion: users may directly manipulate object paths in 3D through intuitive interactive editing, or simply supply a desired camera trajectory to guide viewpoint changes. Together, these components enable SymphoMotion to generate videos that faithfully follow user-defined camera motion and object dynamics within a unified, coherent framework.

A further challenge lies in the lack of real-world datasets that jointly annotate camera and object motion. Existing datasets usually cover only one modality: camera-centric datasets such as RealEstate10K~\cite{zhou2018stereo} and ACID~\cite{xiao2021development} provide diverse camera trajectories but mostly depict static scenes, while object-centric datasets such as MagicData~\cite{li2025magicmotion} and 360°-Motion~\cite{fu20243dtrajmaster} capture rich object motion but assume a fixed or nearly fixed camera. Although synthetic datasets like SynFMC~\cite{shuai2025free} and OmniWorld-Game~\cite{zhou2025omniworld} include both types of motion, the domain gap limits their applicability to real-world video generation.
To fill this gap, we introduce RealCOD-25K, a large-scale real-world dataset with paired annotations of camera poses and object-level 3D trajectories. RealCOD-25K contains more than 25K video clips spanning diverse indoor and outdoor environments, each sequence providing synchronized camera motion and 3D object trajectories. This comprehensive dataset offers the necessary supervision for learning unified camera–object motion and serves as a robust benchmark for evaluating systems such as SymphoMotion.

In summary, our main contributions are threefold:
\begin{itemize}
  \item 
  We propose SymphoMotion, a unified framework that jointly controls camera motion and object dynamics within a single model, enabling coherent, flexible, and depth-aware motion specification that remains consistent across diverse viewpoints and scene configurations.
  \item We introduce RealCOD-25K, 
  a comprehensive real-world dataset providing paired annotations of camera poses and object-level 3D trajectories across diverse scenes, addressing a critical data gap for training and evaluating unified motion-control models.
  \item Extensive experiments and user studies show that SymphoMotion outperforms state-of-the-art methods in both visual fidelity and motion controllability.
\end{itemize}
\section{Related Work}
\label{sec:Related}

\noindent \textbf{Camera Controlled Video Diffusion Models.~}
To enable camera pose control in video generation, CameraCtrl~\cite{he2024cameractrl} and I2VControl-Camera~\cite{feng2024i2vcontrol} inject camera parameters, such as Plücker embeddings~\cite{sitzmann2021light} or point trajectories, into pretrained video diffusion models.
Building on these methods, CamCo~\cite{xu2024camco} incorporates epipolar geometry into attention layers to preserve multi-view consistency, while CamTrol~\cite{hou2024training} uses 3D point clouds to improve geometric awareness.
AC3D~\cite{bahmani2025ac3d} further refines camera-representation injection, while Uni3C~\cite{cao2025uni3c} and VD3D~\cite{bahmani2024vd3d} extend camera control to transformer-based video diffusion architectures~\cite{menapace2024snap}.
Beyond single-camera settings, CVD~\cite{kuang2024collaborative} and SyncCamMaster~\cite{bai2024syncammaster} support multi-camera synchronization and cross-view video generation. In addition, CameraCtrl II~\cite{he2025cameractrl} enables camera-controlled dynamic scene synthesis with a dedicated dataset.
Despite this progress, these methods are limited to camera control and cannot manipulate object dynamics. In contrast, SymphoMotion enables controllable camera motion and dynamic object manipulation.

\vspace{3mm}
\noindent \textbf{Object Controlled Video Diffusion Models.~}
Object-controllable video generation has recently attracted attention for enabling precise object control during video synthesis.
Early approaches, including Motion-i2V~\cite{shi2024motion}, DragNUWA~\cite{yin2023dragnuwa}, and Tora~\cite{zhang2025tora}, incorporate optical flow into video generation frameworks to control object motion.
Building on point-map guidance, TrackGo~\cite{zhou2025trackgo} represents objects with key points and integrates them through a custom adapter.
Other methods, such as Peekaboo~\cite{jain2024peekaboo}, MagicMotion~\cite{li2025magicmotion}, and Boximator~\cite{wang2024boximator}, use 2D bounding boxes as explicit spatial priors for trajectory control. By encoding box coordinates into the diffusion process, they constrain object positions and scales across frames, enabling effective trajectory supervision.
Inspired by GLIGEN~\cite{li2023gligen}, the training-free framework FreeTraj~\cite{qiu2024freetraj} incorporates bounding-box conditioning into video diffusion models by modifying attention layers or the initial noised video latents.
Several studies~\cite{fu20243dtrajmaster,wang2025levitor} further explore 3D trajectory-based control for more sophisticated motion synthesis. LeViTor~\cite{wang2025levitor} uses depth-augmented keypoint trajectory maps to capture spatial structure, while 3DTrajMaster~\cite{fu20243dtrajmaster} designs customized 3D trajectories to model object motion.
However, these methods focus on object motion while largely neglecting camera movement. In contrast, SymphoMotion jointly controls camera motion and object dynamics through dedicated mechanisms, providing unified fine-grained spatiotemporal control over video generation.

\vspace{1mm}
\noindent \textbf{Camera and Object Controlled Video Diffusion Models.~}
Recent studies have advanced motion control by jointly modeling camera and object motion, representing a major step toward unified camera and object controlled video generation. MotionPrompting~\cite{geng2025motion}, ImageConductor~\cite{li2025image}, and ATI~\cite{wang2025ati} define motion priors through optical flow, 2D point tracking, or feature similarity, enabling users to interactively control both camera and object motion. However, this coupled camera–object control paradigm is effective only in scenarios involving limited motion amplitudes, where both camera and object dynamics remain relatively constrained.
Several recent methods have further decoupled and refined the control of camera and object dynamics.
Perception-as-Control~\cite{chen2025perception} trains separate modules for camera and object motion, enabling each to be optimized independently.
VidCraft3~\cite{zheng2025vidcraft3} proposes a disentangled control framework spanning multiple motion modalities, enabling coordinated motion generation.
MotionCtrl~\cite{wang2024motionctrl} injects extrinsic matrices into diffusion models to achieve camera pose control, while simultaneously processing point maps with a Gaussian filter and trainable encoders to represent object trajectories.
Nevertheless, these approaches restrict object control to the two-dimensional plane, resulting in suboptimal motion control.
To address this limitation, FMC~\cite{shuai2025free} employs 6D pose representations to more accurately capture object motion in three-dimensional space.
Although this formulation improves geometric fidelity, the performance of FMC remains limited by its reliance on a synthetic dataset.
Furthermore, its requirement for explicit 6D pose inputs increases operational complexity and reduces user intuitiveness, limiting its practicality in real-world scenarios.
Compared with previous approaches, SymphoMotion, trained on the RealCOD-25K, provides a unified and flexible framework that not only enables precise and controllable camera motion in video generation but also supports more realistic and spatially-consistent object manipulation.

\begin{figure*}[t]
\centering
\includegraphics[width=0.99\textwidth]{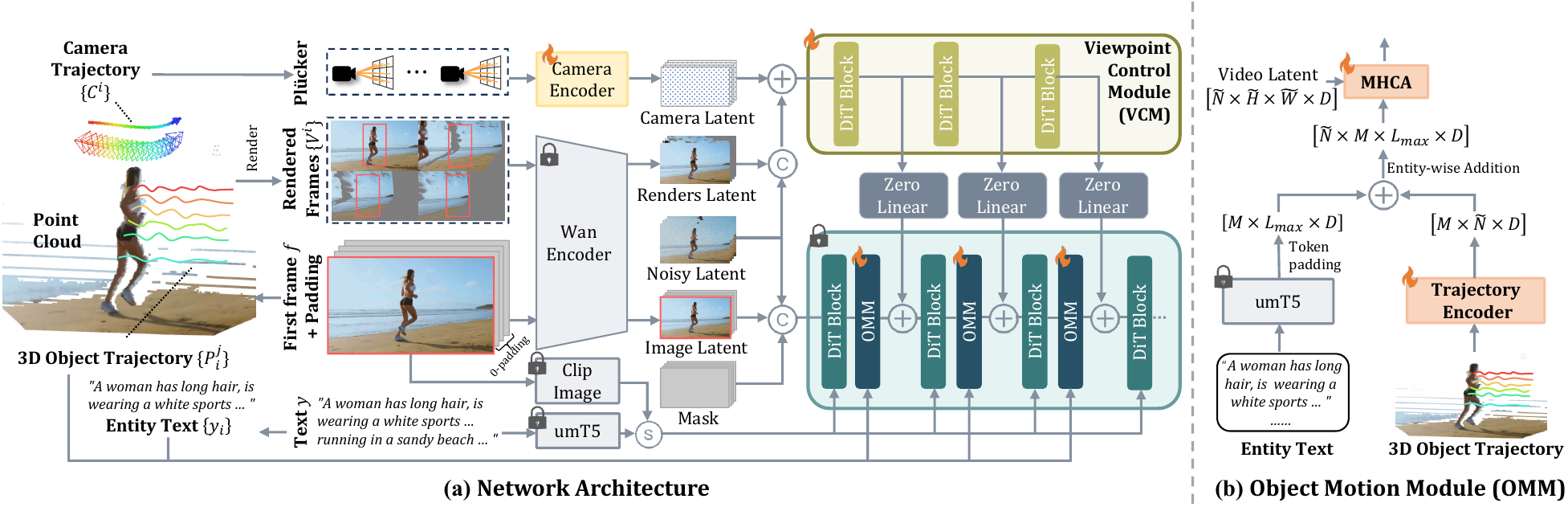}
\vspace{-3mm}
\caption{
{
\textbf{Overview of SymphoMotion.} 
Built on Wan-I2V~\cite{wang2025wan}, 
SymphoMotion introduces two complementary mechanisms for simultaneous control of camera and object motion: Camera Trajectory Control (CTC) and Object Dynamics Control (ODC).
Given a reference image, a text prompt, and the specified camera and object trajectories, CTC employs the Viewpoint Control Module (VCM) to integrate 3D geometric priors with camera motion for precise camera trajectory control. In parallel, ODC, powered by the Object Motion Module (OMM), combines 2D visual guidance with 3D motion cues to achieve dynamic and spatially coherent object manipulation.}
}
\vspace{-2.5mm}
\label{fig:method}
\end{figure*}

\section{SymphoMotion}
\label{sec:Method}

Controlling both camera and object motion in video generation remains challenging, as it requires precise coordination of global and local spatial dynamics in 3D space. To tackle this challenge, we propose SymphoMotion, a diffusion-based framework that enables synchronized and disentangled control over 3D-aware camera and object motions. As shown in Fig.~\ref{fig:method}, the input to our method includes: a reference image $f\in\mathbb{R}^{3\times h\times w}$, a set of camera trajectories $\{C^i\}_{i=1}^N$ specifying the viewpoints of $N$ target frames, a text prompt $y$ consisting of $M$ moving objects $\{y_{i}\}_{i=1}^M$, and their associated 3D motion trajectories $\{ P_{i}^{j} \}_{i=1,\; j=1}^{M,\; N}$. Our framework introduces two mechanisms for motion control:
(1) Camera Trajectory Control (CTC) which integrates 3D geometric priors for precise camera control (Section~\ref{sec:CMM}); and
(2) Object Dynamics Control (ODC) that exploits both 2D and 3D spatial cues to model realistic object motion (Section~\ref{sec:OMM}).
Training details are provided in Section~\ref{sec:Training}, and the inference pipeline is described in Section~\ref{sec:Inference}, following a review of the base diffusion model in Section~\ref{sec:perliminary}.

\subsection{Preliminary}
\label{sec:perliminary}
\textbf{Video Diffusion Models.}
Latent diffusion models perform the denoising process in a learned latent space rather than directly in pixel space, significantly improving both efficiency and scalability~\cite{rombach2022high}. 
Given a training video $x$, we employ a pre-trained 3D variational autoencoder to encode it into a latent representation $z_0$. 
The forward process gradually adds Gaussian noise $\epsilon \sim \mathcal{N}(0, I)$ to the latent variable over $T$ timesteps, generating intermediate noisy latent $z_t$ for $t \in [0, T]$. 
The training objective is to optimize a denoising network $\epsilon_{\theta}$ to predict the added noise:
\begin{equation}
\label{eq:diffusion}
\min_\theta \mathbb{E}_{z_0, t, \epsilon, c_{y}, c_{f}}
\left[\|\epsilon_\theta(z_t, t, c_{y}, c_{f}) - \epsilon\|^2\right],
\end{equation}
where $c_{y}$ and $c_{f}$ denote the conditioning embeddings extracted from text $y$ and image $f$, respectively.
Recently, most video diffusion models have employed Flow Matching~\cite{lipman2022flow} as an improved diffusion formulation, offering faster convergence and more stable training. 
Based on the ordinary differential equations, Flow Matching defines the linear interpolation between $z_0$ and $z_1$:
\begin{equation}
z_t = t z_1 + (1 - t) z_0,
\end{equation}
where $t\in[0,1]$ is sampled from the logit-normal distribution.
The ground-truth velocity is defined as $v_t = \frac{dz_t}{dt} = z_1 - z_0$, and the model is trained to predict it by minimizing: 
\begin{equation}
\label{eq:flow_matching}
\min_\theta \mathbb{E}_{z_0, t, \epsilon, c_y, c_f}
\left[\|v_\theta(z_t, t, c_y, c_f) - v_t\|^2\right].
\end{equation}

\noindent\textbf{Diffusion Transformer (DiT).}
Recent work has explored transformer architectures for diffusion models in lieu of the traditional UNet backbone~\cite{ronneberger2015u}, which better capture long-range temporal dependencies. The Diffusion Transformer (DiT)~\cite{peebles2023scalable} adopts self-attention over spatio-temporal tokens to enhance video coherence and quality. We build upon Wan-I2V~\cite{wang2025wan}, which injects textual features from the multi-language encoder umT5~\cite{chung2023unimax} into Wan-I2V through cross-attention and incorporates visual features from CLIP’s image encoder~\cite{radford2021learning} to enhance image-to-video synthesis.

\subsection{Camera Trajectory Control}
\label{sec:CMM}
\noindent\textbf{3D Geometric Priors.}
Previous camera-controlled video generation methods typically encode camera embeddings using Pl{\"u}cker rays~\cite{bahmani2024ac3d, liang2024wonderland}. 
However, such representations only capture the camera pose, lacking rich structural information about the underlying 3D scene. This makes it challenging to maintain geometric consistency. To address this, we draw inspiration from ViewCrafter~\cite{yu2024viewcrafter} and introduce point clouds as 3D geometric priors, providing complementary structural cues that enhance spatial coherence and geometric fidelity during camera control.
Specifically, given a reference image $f$, we estimate its point cloud, camera intrinsics, and pose $C^f$ using Depth-Pro~\cite{bochkovskii2024depth}. 
The camera is then navigated along a target pose sequence $\mathcal{C} = \{C^1, \ldots, C^N\}$, 
with $C^1$ aligned to the reference pose $C^f$.
By rendering the point cloud from these viewpoints, we obtain a set of geometry-aware frames 
$\mathcal{V} = \{V^{1}, \ldots, V^{N}\}$, where $V^{1}$ corresponds to the reference image $f$.

\vspace{1mm}
\noindent\textbf{Camera Motion Injection.} 
As illustrated in Fig.~\ref{fig:method}, we employ two encoders to capture camera motion and geometric context: a camera encoder and a Wan encoder. Their outputs are fused and fed into the Viewpoint Control Module (VCM), which injects camera motion into the video generation model to enable precise and controllable viewpoint transitions.
Specifically, the camera encoder processes Pl{\"u}cker embeddings of the target pose sequence $\mathcal{C}$ to produce the motion representation $c_{{cam}}$, 
using a canonical pose (with zero translation) for the first frame and relative poses for subsequent ones. 
In parallel, the Wan encoder extracts geometry-aware features $c_{{pcd}}$ from the rendered point-cloud frames, capturing both 3D structure and visual context.
To fuse the two, $c_{{pcd}}$ is first concatenated with the noisy latent $z_t$ to enrich geometric awareness. Then, $c_{{cam}}$ is added to provide explicit motion cues. The resulting unified representation is passed to the VCM, which is implemented as a ControlNet $\phi_{\theta}$.
The training objective is:
\begin{equation}
\label{eq:diffusion-cam-motion}
\begin{aligned}
    &\hspace{3em}
    \min_\theta \; \mathbb{E}_{z_0, t, \epsilon, c_{y}, c_{f}, c_{cam}, c_{pcd}} \\
    &\left[\|v_\theta(z_t, t, c_{y}, c_{f},
    \phi_{\theta}(c_{cam}, c_{pcd})) - v_t\|^2\right].
\end{aligned}
\end{equation}

\subsection{Object Dynamics Control}
\label{sec:OMM}
To achieve fine-grained control over object motion, we introduce a dedicated object dynamics control mechanism, guiding object movement using 3D object trajectories $\{ P_{i}^{j} \}_{i=1,\; j=1}^{M,\; N}$, where $P_i \in \mathbb{R}^{N \times N_p \times 3}$ denotes the positions of $N_p$ points sampled from the $i$-th object over $N$ frames.
To enable accurate motion behavior, we leverage both 2D visual guidance and 3D motion information. 
The 2D guidance establishes explicit spatial anchors in the image plane, 
constraining object localization to follow the predefined visual trajectory.
Meanwhile, the 3D motion trajectories provide geometry-aware supervision, maintaining coherent spatial relationships across views. 
Together, these complementary signals enable reliable object motion.

\vspace{1mm}
\noindent\textbf{2D Visual Guidance.}
For each moving object $i$, we derive a 2D trajectory ${P_{\text{2D}}}_i$ by projecting its 3D trajectory $P_i$ 
onto the image plane using the target camera poses $\mathcal{C}$. 
Based on the projected points ${P_{\text{2D}}}_i$, we fit per-frame bounding boxes that delineate the object’s expected position in pixel coordinates. 
As illustrated in Fig.~\ref{fig:method}, these bounding boxes are directly rendered onto the point-cloud frames $\mathcal{V}$, serving as explicit spatial anchors that guide the model in localizing each object across frames.
By overlaying motion boxes in the rendered input rather than encoding solely in latent space, we provide the model with strong visual cues to track the image-plane projection of each object’s 3D motion path.

\vspace{1mm}
\noindent\textbf{3D Trajectory Conditioning.}
We further provide the 3D motion cues in object dynamics control via an Object Motion Module (OMM). As shown in Fig.~\ref{fig:method} (b), each object’s 3D trajectory $P_i$ is first transformed into the coordinate system of the reference camera $C^f$, and subsequently encoded into latent embeddings using a trajectory encoder composed of a linear projection layer and a temporal downsampler ($N$ frames to $\tilde{N}$ frames). 
Meanwhile, the entity prompts $y_{i}$ are converted into semantic embeddings using the frozen language encoder.
The two embeddings are fused through element-wise addition to produce motion-aware representations $c_{obj}$. The whole process can be formulated as:
\begin{equation}
c_{obj}
= \psi_{\theta}\!\big(
\{ P_{i}^{j} \}_{i=1,\; j=1}^{M,\; N}\;, \{y_{i}\}_{i=1}^M
\big),
\end{equation}
where $\psi_{\theta}$ represents the fusion pipeline for objects' 3D motion trajectory and semantic identity.
We then integrate $c_{obj}$ into the diffusion model by modifying each transformer block to cross-attend to it:
\begin{equation}
\begin{array}{l}
Z'_i = Z_i + \mathrm{CrossAttn}\!\Big(
Q = Z_i,
K = c_{obj},
V = c_{obj}
\Big),
\end{array}
\end{equation}
where $Z_i$ denotes the latent features at layer $i$. By attending to the motion-aware tokens, the model aligns its latent representation with the specified 3D trajectories, enabling consistent and controllable object motion during generation. The overall training objective is defined as:
\begin{equation}
\label{eq:diffusion-train}
\begin{aligned}
    &\hspace{3em}
    \min_\theta \; \mathbb{E}_{z_0, t, \epsilon, c_{y}, c_{f}, c_{cam}, c_{pcd}}
    \left[\|v_\theta(z_t, t, c_{y}, c_{f}, \right.\\
    &\left. 
    \phi_{\theta}(c_{cam}, c_{pcd}),
    \psi_{\theta}\!(\{ P_{i}^{j} \}_{i=1,\; j=1}^{M,\; N}\;, \{y_{i}\}_{i=1}^M)) - v_t\|^2\right].
\end{aligned}
\end{equation}

\subsection{Training Strategy}
\label{sec:Training}

\noindent\textbf{Data Construction.}
To enable controllable camera motion and dynamic object manipulation, the training data must include video clips annotated with captions, camera poses, and object trajectories corresponding to given reference images and text prompts.
Since no existing dataset provides such comprehensive annotations, we construct RealCOD-25K dataset, containing 25K high-quality video clips spanning diverse real-world scenes (see Section~\ref{sec:RealCOD-25K} for details).

\vspace{1mm}
\noindent\textbf{Training Procedure.}
SymphoMotion is built upon the pre-trained Wan-I2V~\cite{wang2025wan}, which remains frozen during training.
Training is performed on video sequences of 81 frames at a resolution of 832×480.
Following MotionCtrl~\cite{wang2024motionctrl}, we adopt a two-stage strategy:
(1) The CTC part is first trained to learn camera control;
(2) The CTC is then frozen while the ODC part is trained for object motion control.
All experiments are conducted on 32 NVIDIA H100 GPUs with a total batch size of 32.
We use AdamW~\cite{loshchilov2017decoupled} as the optimizer.
The learning rate is linearly warmed up to $1\times10^{-5}$ over the first 400 steps and kept constant thereafter.

\subsection{Inference Pipeline}
\label{sec:Inference}

\begin{figure}[t]
\centering
\includegraphics[width=0.48\textwidth]{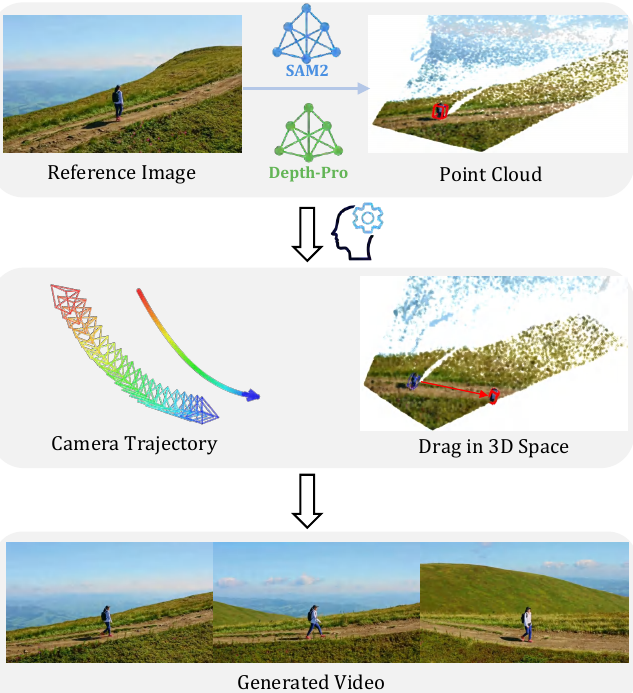}
\vspace{-7mm}
\caption{\textbf{Inference pipeline of SymphoMotion.} Users can specify camera motion and interactively draw 3D trajectories of selected objects through our interface, and the system generates videos that align with the user-defined camera and object motion.}
\vspace{-3mm}
\label{fig:inference}
\end{figure}

We design an intuitive interactive system for inference, as shown in Fig.~\ref{fig:inference}. Given a single reference image, the system first reconstructs a dense point cloud using Depth-Pro~\cite{bochkovskii2024depth}. Users select objects via SAM2~\cite{ravi2024sam}; the selected 2D masks are lifted into 3D by projecting pixels onto the reconstructed point cloud, from which an initial 3D bounding box is fitted. Through an interactive panel (detailed in supplementary materials), users can drag and adjust this box in 3D space, and the system records the manipulated box positions as the object’s 3D motion trajectory. Users may simultaneously specify a camera path by defining poses relative to the reference camera. Given both object trajectories and camera motion, SymphoMotion generates a video that follows the user-defined 3D object dynamics and camera movement.

\section{RealCOD-25K Dataset}
\label{sec:RealCOD-25K}

\begin{figure}[t]
\centering
\includegraphics[width=0.47\textwidth]{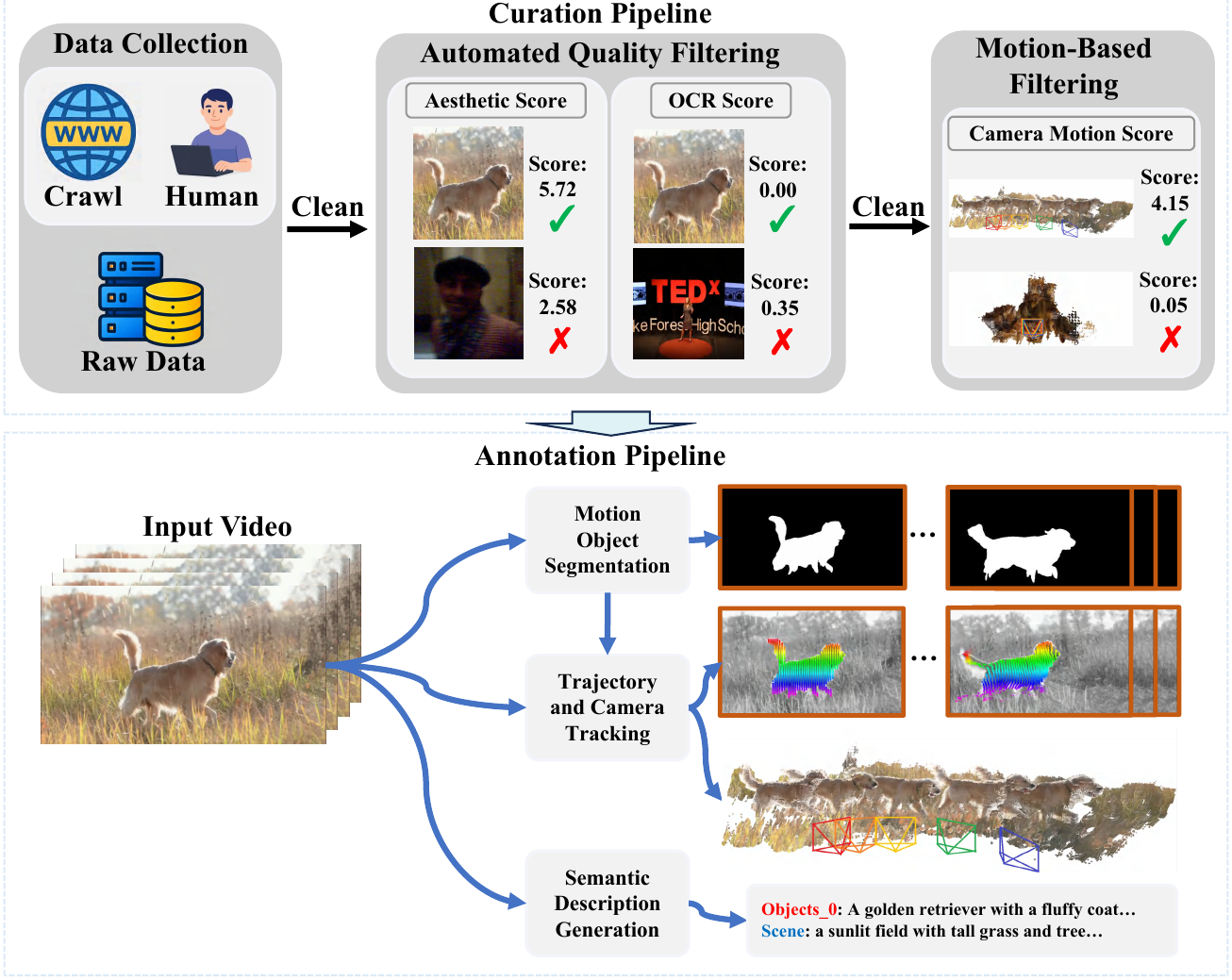}
\vspace{-3mm}
\caption{\textbf{RealCOD-25K dataset construction pipeline.}
}
\vspace{-3mm}
\label{fig:data_creation}
\end{figure}

To support large-scale training and unified evaluation, we construct RealCOD-25K, a curated dataset tailored for controllable camera and object dynamics. As shown in Figure~\ref{fig:data_creation}, RealCOD-25K is built through two pipelines.

\subsection{Curation Pipeline}

\noindent\textbf{(1) Data Collection.}
We collected one million real-world video clips through a combination of automated web crawling and manual curation from publicly available platforms such as YouTube, Mixkit, Pexels, and Pixabay, covering diverse scenes with rich camera and object motions.

\noindent\textbf{(2) Automated Quality Filtering.}
From the initial one million videos, we automatically filtered low-quality samples using the LAION aesthetic predictor and PaddleOCR~\cite{liao2022real}.
The aesthetic predictor retained videos with scores above 5, while PaddleOCR reliably removed clips containing visibly excessive overlaid text, such as watermarks or subtitles. Extremely short or corrupted clips were further discarded.

\noindent\textbf{(3) Motion-Based Filtering.}
To ensure stable and reliable geometry information annotation in later stages, our motion filtering pipeline leverages the lightweight VMAF metric~\cite{li2016toward} to retain videos with sufficient motion diversity. In parallel, to guarantee adequate foreground dynamics and realistic motion patterns, the vision–language model Qwen-2.5-VL-72B~\cite{bai2025qwen2} removed videos lacking moving objects. This filtering reduced the dataset to approximately 35K clips exhibiting both camera and object motion.

\noindent\textbf{(4) Manual Curation and Finalization.}
Five researchers with expertise in computer vision manually reviewed the remaining videos over 120 person-hours to identify and remove residual low-quality samples. The manual inspection emphasized visual fidelity, motion consistency, and geometric plausibility, resulting in the RealCOD-25K dataset comprising 25K high-quality clips that consistently exhibit meaningful camera motion and coherent foreground object dynamics across diverse real-world scenes.

\subsection{Annotation Pipeline}

\noindent\textbf{(1) Motion Object Segmentation.}
To annotate moving objects, we first employed SegAnyMo~\cite{huang2025segment}, which takes a video as input and effectively predicts segmentation masks for all moving foreground objects. For each distinct object, the model generates an initial mask on the first frame and assigns a unique identifier, enabling consistent object identity tracking across subsequent frames.

\noindent\textbf{(2) Camera and Trajectory Tracking.}
For geometric estimation, we adopt MegaSAM~\cite{li2025megasam} as the base pipeline and replace its original monocular and metric depth modules with Depth Anything V2~\cite{yang2024depth} and UniDepth V2~\cite{piccinelli2025unidepthv2}, respectively, to obtain more accurate and temporally consistent depth.
Based on the recovered geometry, we further estimate object motion trajectories. Initialized with first-frame object masks, we apply SpatialTrackerV2~\cite{xiao2025spatialtrackerv2} to track object motion across frames. The tracker estimates per-object 2D trajectories conditioned on the geometry, which are then lifted to 3D space via back-projection.

\noindent\textbf{(3) Semantic Description Generation.}
To provide object-level semantic annotations, we employ the large vision–language model Qwen-2.5-VL-72B~\cite{bai2025qwen2} to generate detailed textual descriptions of each moving object, including its appearance, motion patterns, and surrounding scene context. These captions complement the estimated 3D object trajectories, yielding fine-grained, semantically grounded annotations naturally aligned with object-level dynamics.

\section{Experiment}
\subsection{Evaluation}

\noindent\textbf{Evaluation Datasets.}
In the absence of an existing dataset containing both camera and object motion, we curated a diverse collection of 100 real-world videos from publicly available sources, carefully selected to cover a wide range of camera trajectories and dynamic object movements.

\noindent\textbf{Evaluation Metrics.}
Following prior work~\cite{wang2024motionctrl, li2025magicmotion, fu20243dtrajmaster, shuai2025free}, we evaluate performance across four key dimensions:
(1) Visual Quality. We use Fréchet Image Distance (FID)~\cite{maximilian2020fid} to evaluate visual fidelity and Fréchet Video Distance (FVD)~\cite{unterthiner2018towards} to assess temporal coherence.
(2) Text Alignment. CLIP Similarity (CLIPSIM)~\cite{wu2021godiva} measures the semantic consistency between each generated video and its corresponding text prompt.
(3) Camera Motion. Following CameraCtrl, we adopt CamTransErr and CamRotErr to quantify the translation and rotation deviations between the generated and reference camera trajectories.
(4) Object Motion. We use Box-IoU to evaluate the accuracy of object trajectories. For each generated video, we obtain predicted masks $M_{\text{gen}}$ by providing the ground-truth first-frame masks $M_{\text{gt}}(0)$ to SAM2~\cite{ravi2024sam}. Bounding boxes are then derived from each mask, and the mean Intersection-over-Union (IoU) between predicted and ground-truth boxes across all frames is reported as the final Box-IoU score.

\subsection{Comparisons with State-of-the-Art Methods}
We begin by evaluating SymphoMotion under independent camera control, comparing it against prior approaches~\cite{he2024cameractrl,yu2024viewcrafter,cao2025uni3c}. We then assess its capability for unified control, where both motions must be coordinated within a single framework. Across both settings, SymphoMotion consistently demonstrates superior controllability and fidelity.

\begin{figure}
    \centering
    \includegraphics[width=0.999\linewidth]{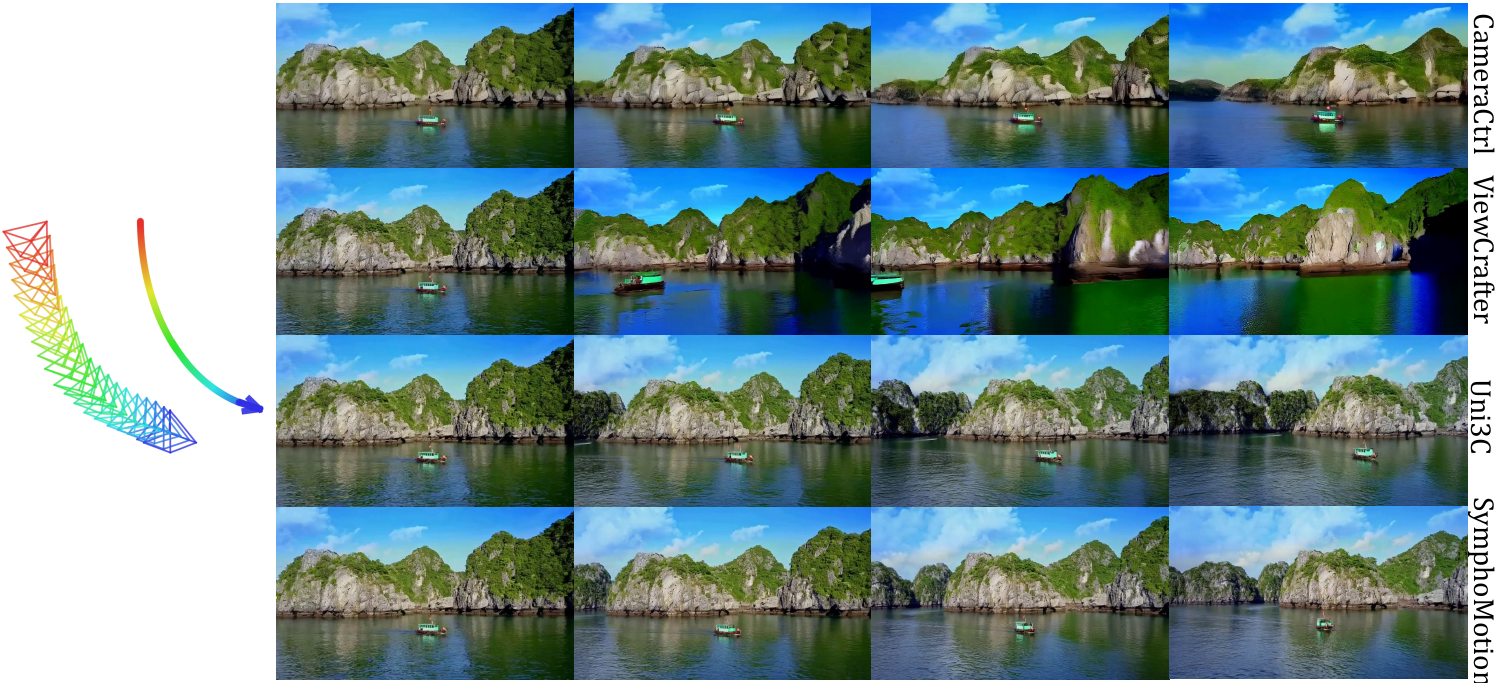}
    \vspace{-7mm}
    \caption{\textbf{Independent camera motion control.}}
    \label{fig:camera-control}
\end{figure}

\noindent\textbf{Independent Control of Camera Motion.}
CameraCtrl~\cite{he2024cameractrl}, ViewCrafter~\cite{yu2024viewcrafter}, and Uni3C~\cite{cao2025uni3c} are selected for comparison, as all accept explicit camera specifications. As shown in Fig.~\ref{fig:camera-control}, when simulating controlled camera motion, CameraCtrl produces noticeable distortions, while ViewCrafter fails to preserve static objects during viewpoint changes. In contrast, Uni3C and SymphoMotion more faithfully reflect the intended camera behavior. The CamTransErr and CamRotErr metrics in Tab.~\ref{tab:performance_metrics} further indicate that SymphoMotion achieves camera-control accuracy comparable to these dedicated baselines.

\noindent\textbf{Simultaneous Control of Camera and Object Motions.}
\begin{figure}[t]
    \centering
    \includegraphics[width=0.97\linewidth]{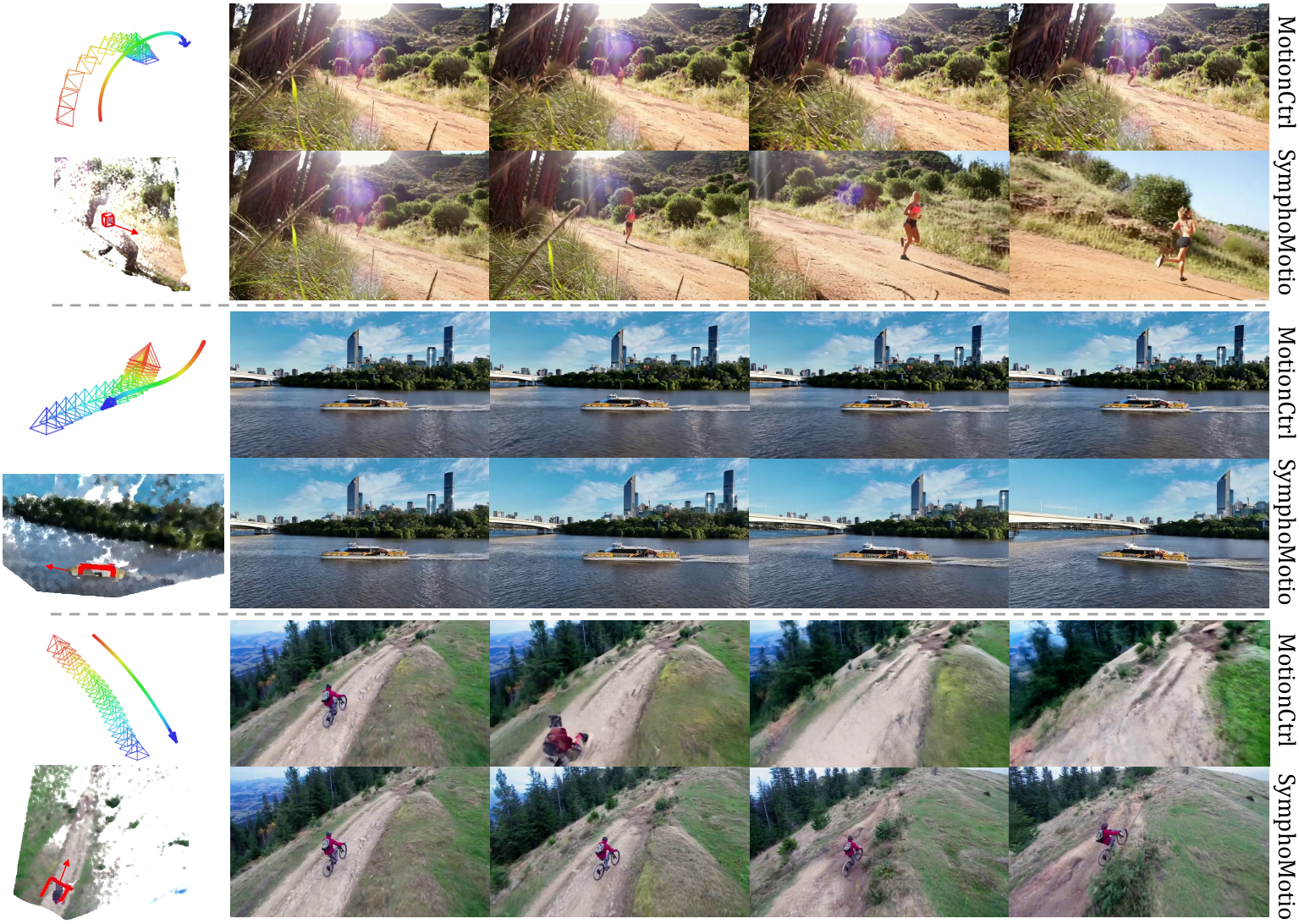}
    \caption{\textbf{Simultaneous control over camera and object motions.} MotionCtrl struggles to generate realistic object dynamics, causing objects to disappear from view, whereas SymphoMotion achieves high-quality simultaneous control.}
    \label{fig:simul_comparison}
\end{figure}
We evaluate joint control of camera and object motion by comparing our method with existing approaches. Since CameraCtrl, Uni3C, and ViewCrafter support only camera motion, we adopt MotionCtrl~\cite{wang2024motionctrl} as the primary baseline for object motion. As shown in Fig.~\ref{fig:simul_comparison}, videos generated by SymphoMotion align more faithfully with the specified control signals, producing coordinated camera movement and realistic object dynamics. In contrast, MotionCtrl exhibits limited controllability for both camera and object motion; its camera trajectories deviate from the prescribed path, and its object behavior remains inconsistent or implausible. Quantitatively, SymphoMotion attains higher Box-IoU scores in Tab.~\ref{tab:performance_metrics}, indicating more accurate object trajectory adherence. Furthermore, as shown in Tab.~\ref{tab:user_study}, SymphoMotion receives higher ratings in the user study, outperforming previous methods in visual and motion quality.

\begin{table}[t]
\centering
\caption{Quantitative comparison of our method SymphoMotion with CameraCtrl, ViewCrafter, Uni3C and MotionCtrl.}
\label{tab:performance_metrics}
\vspace{-3mm}
\scriptsize
\setlength{\tabcolsep}{0.9mm}
\begin{tabular}{l|c|c|c|c|c}
\specialrule{.1em}{.05em}{.05em}
\rowcolor{mygray!50}
Method & CameraCtrl & ViewCrafter & Uni3C & MotionCtrl & SymphoMotion \\
\hline\hline

FID $\downarrow$
& 196.84
& 303.83
& \underline{86.66}
& 182.15
& \textbf{70.47} \\

\rowcolor{mygray2!36}
FVD $\downarrow$
& 1019.49
& 1690.73
& \underline{404.21}
& 738.41
& \textbf{332.50} \\

CLIPSIM $\uparrow$
& 0.29
& 0.28
& \underline{0.31}
& 0.30
& \textbf{0.31} \\

\rowcolor{mygray2!36}
CamTransErr $\downarrow$
& 0.68
& 0.80
& \underline{0.44}
& 0.83
& \textbf{0.37} \\

CamRotErr $\downarrow$
& 0.12
& 0.21
& \underline{0.06}
& 0.23
& \textbf{0.05} \\

\rowcolor{mygray2!36}
Box-IoU $\uparrow$
& --
& --
& --
& \underline{31.42}
& \textbf{61.88} \\
\specialrule{.1em}{.05em}{.05em}
\end{tabular}
\vspace{-4mm}
\end{table}

\subsection{Ablation Studies}
\begin{table}[t]
\centering
\caption{User study on visual quality, text alignment, camera motion, and object motion (scores range from 1 to 5, higher is better).}
\label{tab:user_study}
\vspace{-3mm}
\footnotesize
\setlength{\tabcolsep}{0.96mm}
\resizebox{0.476\textwidth}{!}{
\begin{tabular}{l|c|c|c|c|c}
\specialrule{.1em}{.05em}{.05em} 
\rowcolor{mygray!50}
Method 
& CameraCtrl
& ViewCrafter
& Uni3C
& MotionCtrl 
& SymphoMotion \\
\hline\hline

Visual Quality 
& 3.43
& 4.03
& 4.24
& 3.53
& \textbf{4.87} \\

\rowcolor{mygray2!36}
Text Alignment
& 3.37
& 3.57
& 3.68
& 3.14
& \textbf{4.02} \\

Camera Motion
& 3.13
& 3.47
& 3.97
& 3.18
& \textbf{4.36} \\

\rowcolor{mygray2!36}
Object Motion
& --
& --
& --
& 2.87
& \textbf{4.58} \\

\specialrule{.1em}{.05em}{.05em} 
\end{tabular}
}
\vspace{-1mm}
\end{table}

\begin{table}[t]
\centering
\caption{Quantitative results in ablation study.}
\label{tab:ablation_quantitative}
\vspace{-3mm}
\footnotesize
\setlength{\tabcolsep}{0.6mm}
\resizebox{0.47\textwidth}{!}{
\begin{tabular}{l|c|c|c|c}
\specialrule{.1em}{.05em}{.05em} 
\rowcolor{mygray!50}Metrics & FVD$\downarrow$ & CamTransErr$\downarrow$ & CamRotErr$\downarrow$ & Box-IoU$\uparrow$ \\
\hline\hline

\rowcolor{mygray2!36}
\textit{w/o} $c_{pcd}$ 
& 330.64 & 0.46 & 0.07 & 56.74 \\

\textit{w/o} 2D boxes  
& 337.14 & \textbf{0.36} & 0.06 & 54.32 \\

\rowcolor{mygray2!36}
\textit{w/o} 3D trajectory 
& 343.80 & \textbf{0.36} & 0.06 & 52.16 \\

SymphoMotion 
& \textbf{332.50} & 0.37 & \textbf{0.05} & \textbf{61.88} \\

\specialrule{.1em}{.05em}{.05em} 
\end{tabular}
}
\vspace{-1mm}
\end{table}

\begin{figure}
    \centering
    \includegraphics[width=0.99\linewidth]{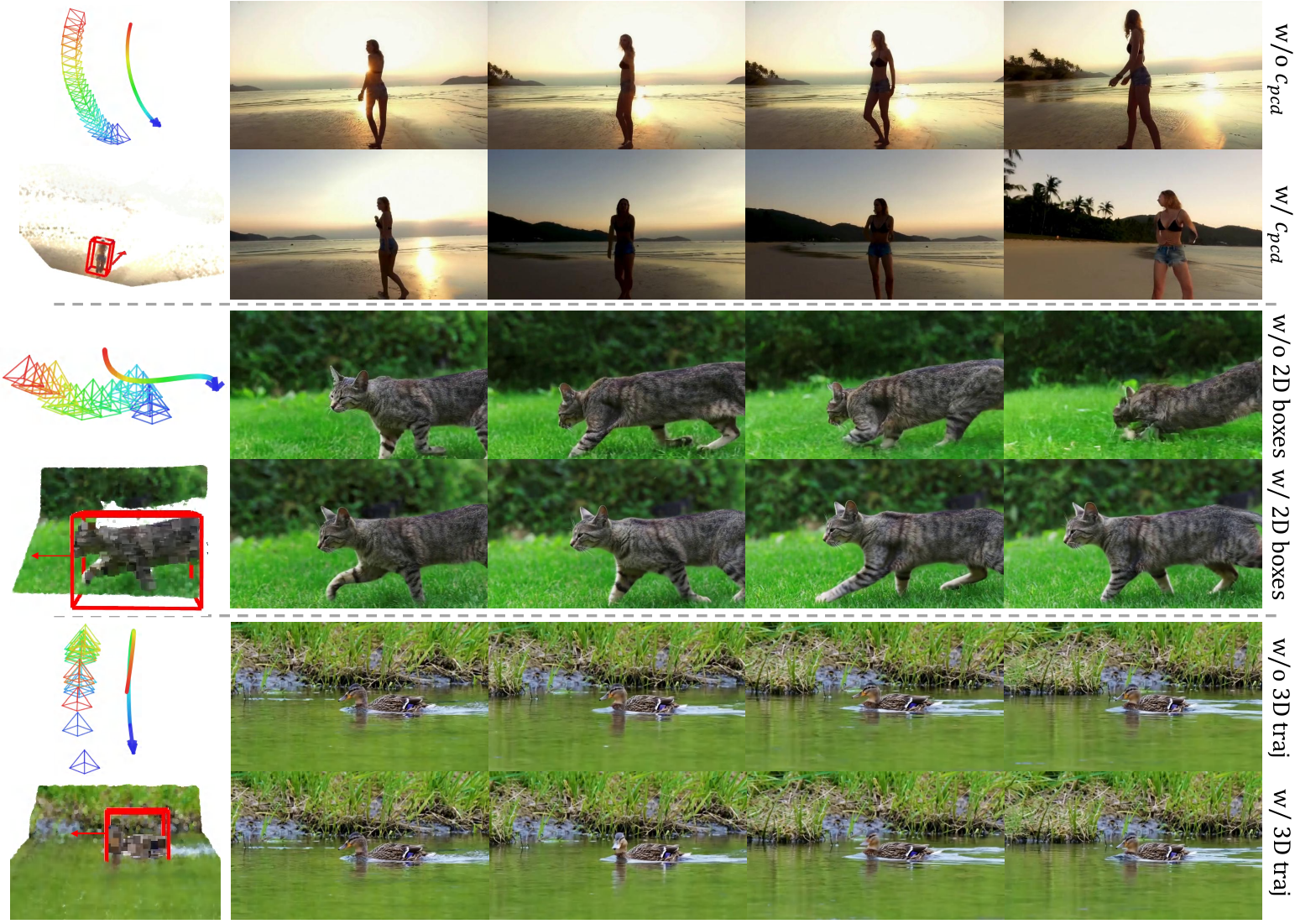}
    \vspace{-2mm}
    \caption{\textbf{Results of different settings in the ablation study.}}
    \label{fig:ablation}
    \vspace{-1mm}
\end{figure}

\noindent\textbf{Effect of 3D Geometric Priors in Camera Trajectory Control.}
As shown in Tab.~\ref{tab:ablation_quantitative} and the first row of Fig.~\ref{fig:ablation}, incorporating geometry-aware renderings yields more accurate camera trajectories due to additional 3D structural priors conveyed by these renderings, which provide richer spatial understanding and informative viewpoint cues. By supplying complementary spatial cues, the geometry-aware renderings help the model better disambiguate camera motion, producing videos that closely follow the target trajectory and exhibit higher geometric consistency.

\vspace{1mm}
\noindent \textbf{Effect of 2D Visual Guidance in Object Dynamics Control.}
2D visual guidance provides an anchor for each object’s projected motion in the image plane. We find that incorporating these cues into the rendered frames enhances training stability and strengthens the model’s ability to reason about object motion. As shown in the second row of Fig.~\ref{fig:ablation}, introducing 2D visual guidance enables the model to better interpret object trajectories, producing videos with more coherent and physically plausible motion.

\vspace{1mm}
\noindent \textbf{Effect of 3D Trajectory Conditioning in Object Dynamics Control.}
3D trajectories provide explicit guidance for modeling object motion in space. As shown in Tab.~\ref{tab:ablation_quantitative}, incorporating object-level 3D trajectories leads to consistently better performance on object-motion metrics. As illustrated in the third row of Fig.~\ref{fig:ablation}, when the model is conditioned on a precise 3D spatial trajectory for the duck, it produces more faithful object motion, such as realistic head turns, indicating a stronger ability to follow complex spatial trajectories.

\section{Conclusion}
This paper presents SymphoMotion, a unified framework for jointly controlling camera motion and object dynamics in video generation. The framework builds on two key mechanisms effectively enabling unified motion control. Camera Trajectory Control leverages pose conditioning with 3D structural information for precise and stable viewpoint manipulation. Object Dynamics Control combines 2D spatial signals with 3D motion representations, allowing objects to follow intended trajectories with accurate motion. Due to the lack of high-quality data, we construct RealCOD-25K to support motion control research. Extensive experiments show that SymphoMotion synthesizes realistic and temporally coherent videos that consistently follow user-specified camera motions and object trajectories.
\clearpage
\setcounter{page}{1}
\twocolumn[{
\renewcommand\twocolumn[1][]{#1}
\maketitlesupplementary
\begin{center}
    \captionsetup{type=figure}
    \includegraphics[width=1\textwidth]{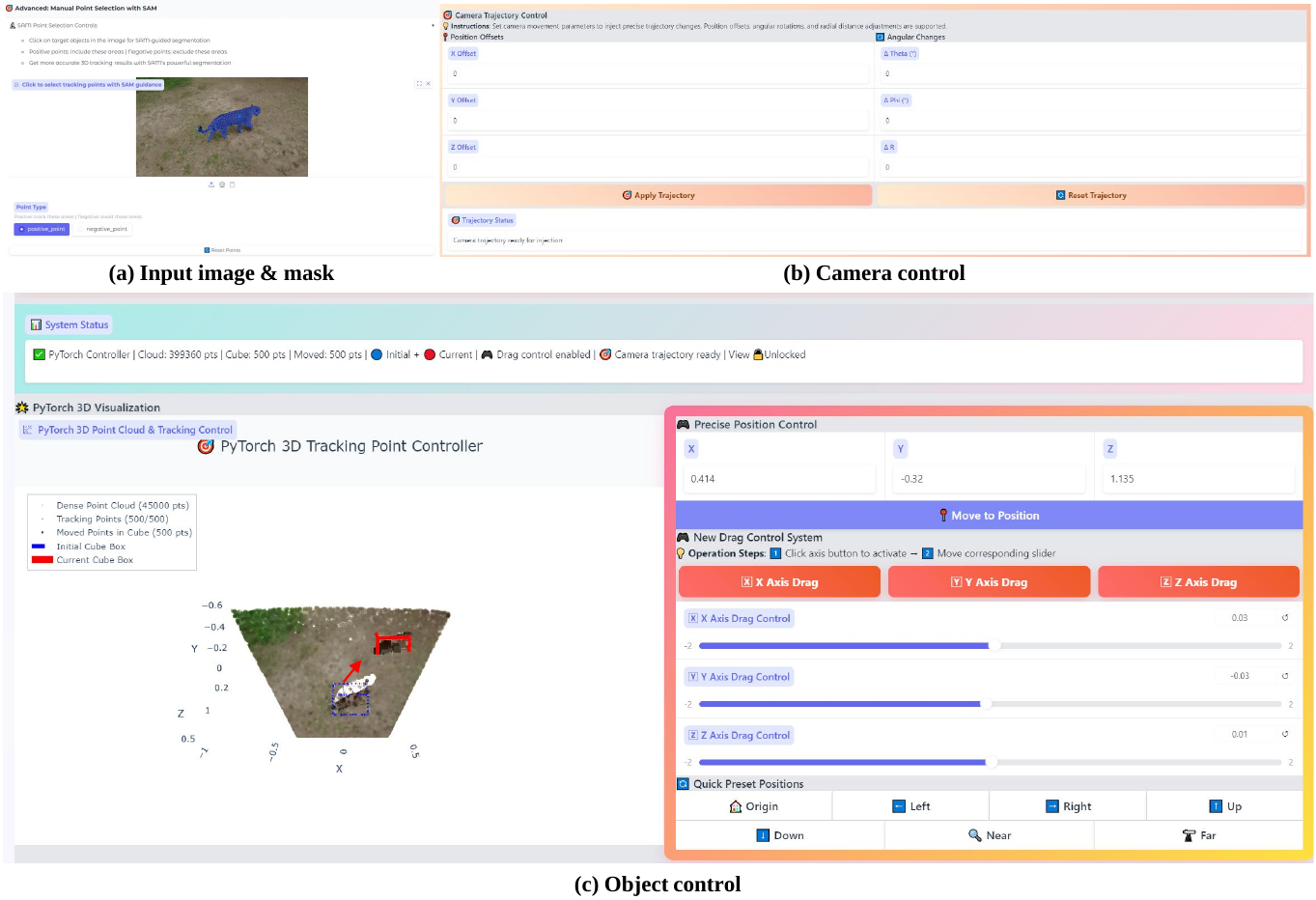}
    \vspace{-8mm}
    \caption{
    \textbf{Interactive Panel Interface.} The panel offers a unified interface for specifying motion inputs to SymphoMotion.
(a) An input image is uploaded, and SAM2 extracts the mask of the target object for subsequent control.
(b) The Camera Control Panel allows users to configure camera movement through rotational and translational adjustments for viewpoint specification.
(c) The Object Control Panel provides interactive editing of 3D object trajectories using the automatically fitted bounding box.
    }
    \label{fig:ui}
\end{center}
}]
\noindent This supplementary material provides inference details and visualization analyses for SymphoMotion, complementing the discussions in the main paper. Specifically, it includes:
\vspace{1mm}
\begin{itemize}
\item Section~\ref{sec:panel}: An overview of the interactive panel used to specify camera and object control signals, along with examples illustrating its use for motion design.
\item Sections~\ref{sec:supplementary-camera}–\ref{sec:supplementary-object}: Additional qualitative comparisons on independent control of camera and object motions.
\item Section~\ref{sec:supplementary-camera-object}: Additional qualitative results on jointly controlling camera motion and object dynamics, showing that SymphoMotion preserves high visual fidelity, accurate camera movement, and reliable object manipulation, whereas prior methods struggle to maintain these properties even under independent camera or object control.
\end{itemize}

\begin{figure}
    \centering
    \includegraphics[width=1\linewidth]{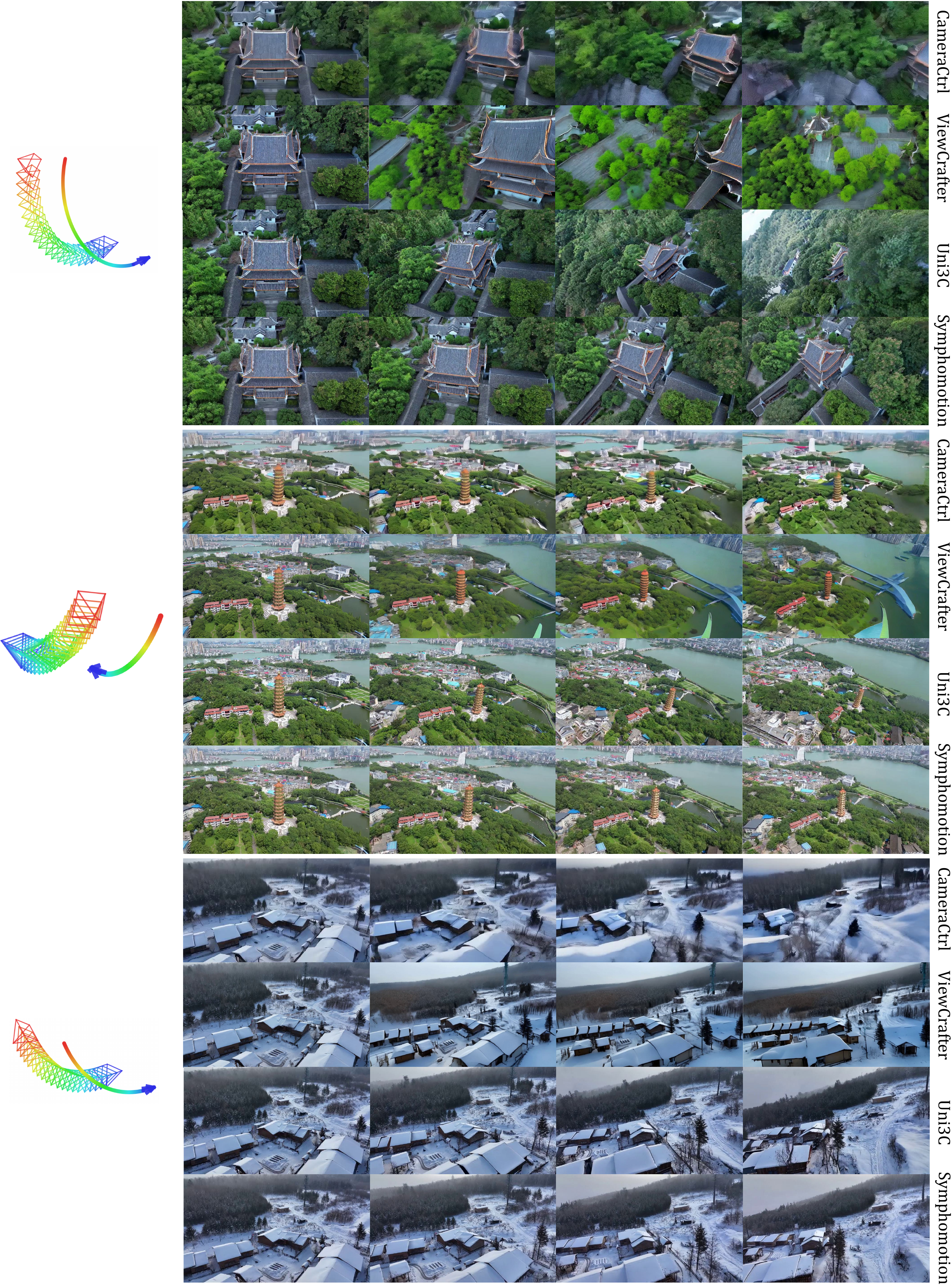}
    \vspace{-2mm}
    \caption{\textbf{Independent camera control for static scene.}}
    \label{fig:material-camera1}
    \vspace{-1mm}
\end{figure}

\section{Interactive Panel}
\label{sec:panel}
The interactive panel provides an intuitive interface for specifying camera movement and object motion before generating videos with SymphoMotion. As shown in Fig.~\ref{fig:ui}, users begin by uploading an input image, from which SAM2 extracts the mask of the object of interest (Fig.~\ref{fig:ui} (a)). The masked region identifies the target object for subsequent control.
After mask selection, users can define motion through two interfaces: the Camera Control Panel (Fig.~\ref{fig:ui} (b)) and the Object Control Panel (Fig.~\ref{fig:ui} (c)). Together, these interfaces offer a simple and flexible way to configure both camera movement and object motion.
\subsection{Camera Control}
\label{sec:panel-camera}
The camera control interface allows users to define the camera movement. As shown in Fig.~\ref{fig:ui} (b), it supports rotational and translational adjustments, including modifications to distance, elevation, azimuth, and spatial offsets. Collectively, these controls specify a camera trajectory that is provided to SymphoMotion as an explicit input for control.

\subsection{Object Control}
\label{sec:panel-object}
To specify object dynamics, the panel incorporates a 3D tracking interface built upon Depth-Pro. After reconstructing a dense point cloud from the first frame, the system automatically fits a 3D bounding box around the selected object. Users can then interactively drag or reposition this bounding box to define the object’s motion trajectory in 3D space. As illustrated in Fig.~\ref{fig:ui}  (c), the blue dashed box denotes the object's initial position, the arrow indicates the user-defined direction of movement, and the red box represents the final position after manipulation.

The resulting trajectory provides a complete 3D motion specification that serves as the object-side control input to SymphoMotion. Combined with the camera trajectory defined in Section~\ref{sec:panel-camera}, this interface enables unified and precise motion inputs for SymphoMotion.

\section{Additional Qualitative Comparisons}
\label{sec:supplementary-comparison}
To further examine motion controllability, we present additional qualitative evaluations across three complementary settings.
Section~\ref{sec:supplementary-camera} investigates independent camera control, assessing how each method follows specified trajectories while maintaining scene stability.
Section~\ref{sec:supplementary-object} analyzes independent object control, focusing on the coherence and consistency of object motion in static environments.
Section~\ref{sec:supplementary-camera-object} evaluates simultaneous control of camera and object motions, examining the capability of each method to coordinate both motion sources within a unified generative framework.
Collectively, these comparisons offer a comprehensive assessment of SymphoMotion's motion controllability across diverse configurations.

\subsection{Independent Control of Camera Motion.}
\label{sec:supplementary-camera}

\begin{figure}
    \centering
    \includegraphics[width=1\linewidth]{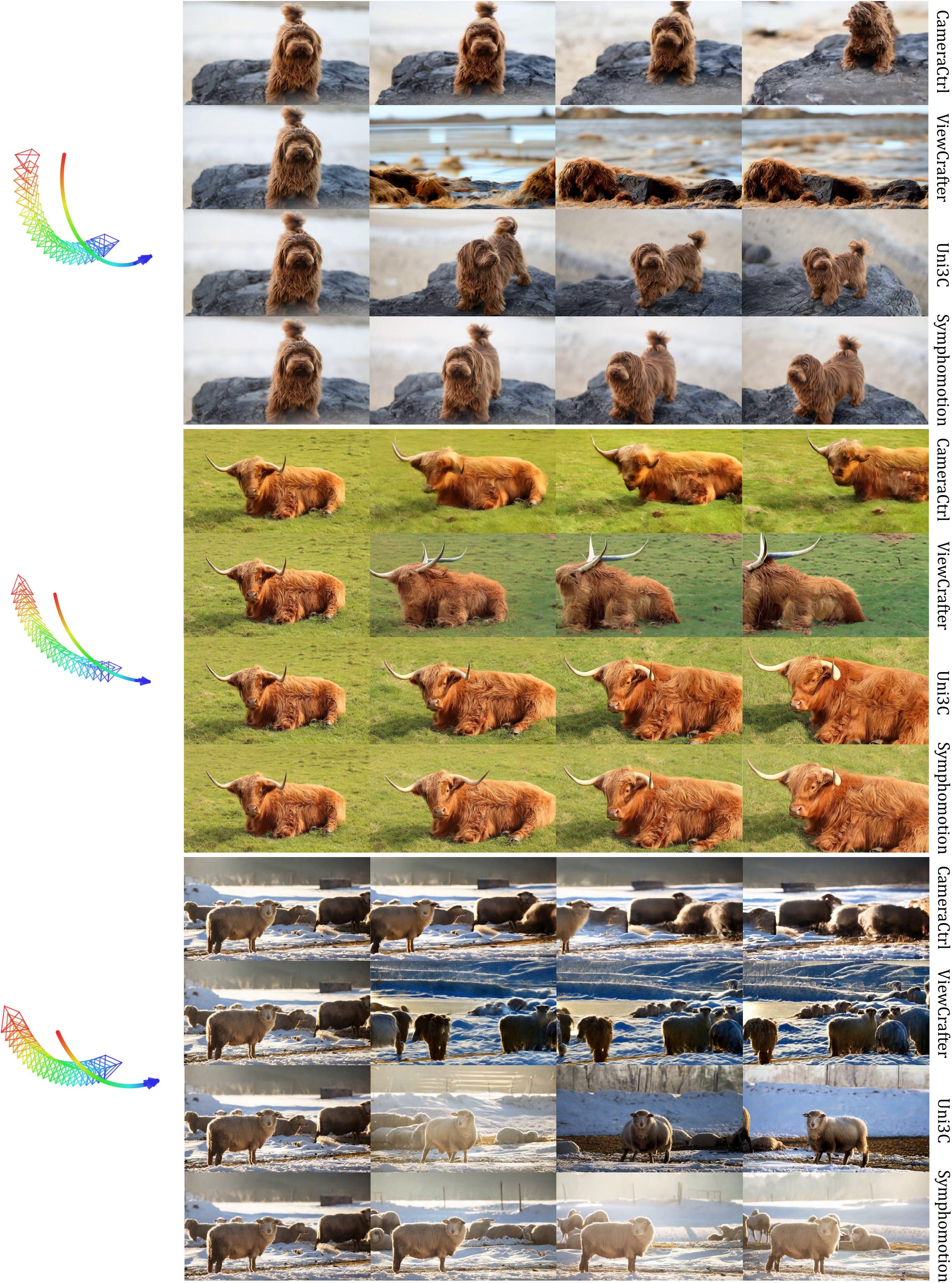}
    \vspace{-2mm}
    \caption{\textbf{Independent camera control for static object.}}
    \label{fig:material-camera2}
    \vspace{-1mm}
\end{figure}

\begin{figure*}[t]
\centering
\includegraphics[width=1\textwidth]{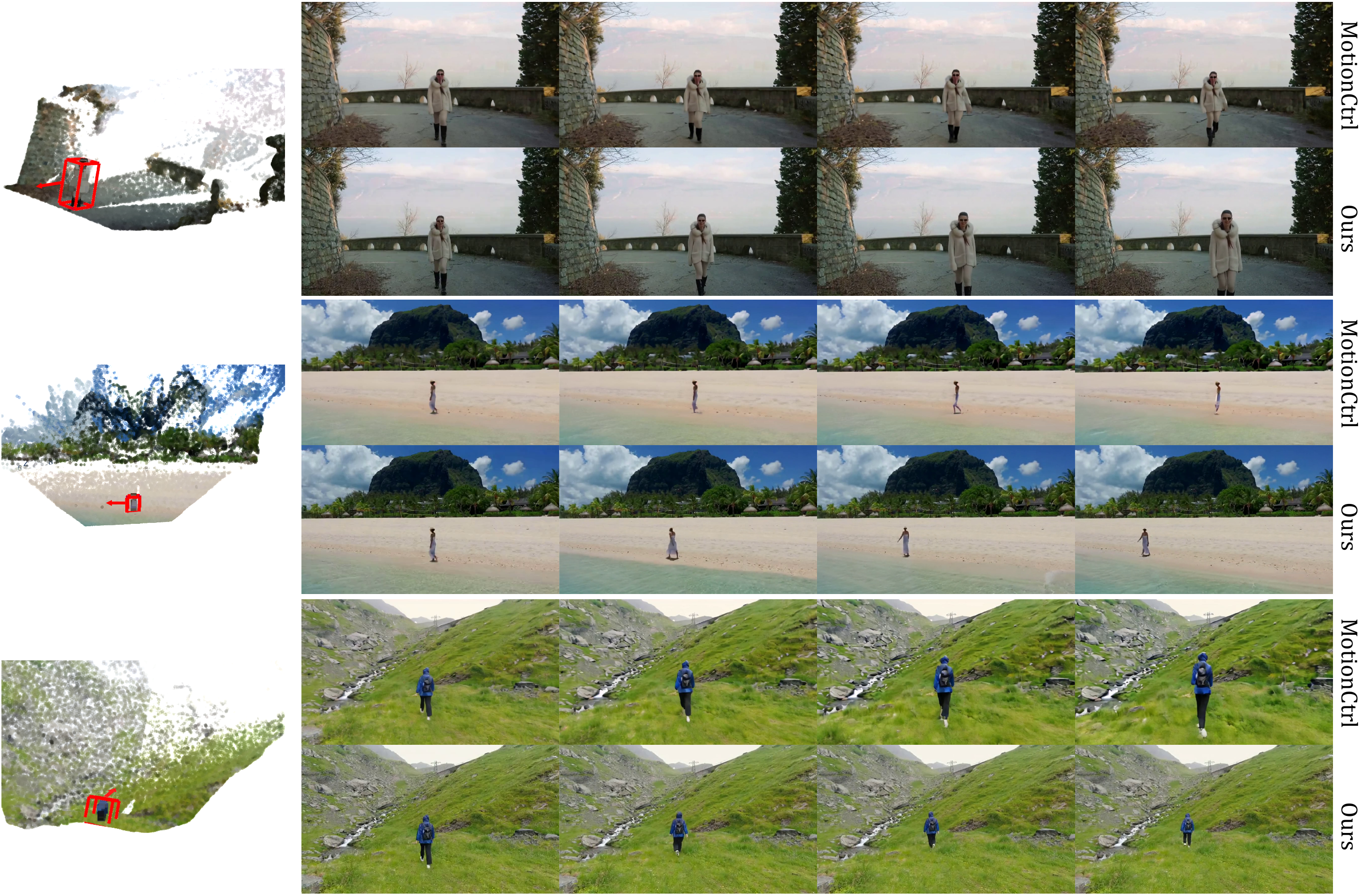}
\caption{
\textbf{More Qualitative Results on Independent Object Motion Control.}
}
\label{fig:material-object}
\end{figure*}
Fig.~\ref{fig:material-camera1} and Fig.~\ref{fig:material-camera2} present additional qualitative comparisons for independent camera control at the scene and object levels, respectively. CameraCtrl consistently shows noticeable drift and reduced video fidelity in both settings, suggesting limited adherence to the intended trajectories and weaker geometric stability under viewpoint changes. ViewCrafter performs reasonably well at the scene level, where camera motion is relatively coarse and global, but deteriorates substantially under object-level cues requiring more precise local control. For instance, in Fig.~\ref{fig:material-camera2}, second row, the target trajectory specifies a zoom-in motion, but ViewCrafter introduces noticeable distortions during camera movement, indicating limited precision in fine-grained camera adjustment and poor consistency under object-centric viewpoint control. In contrast, both Uni3C and SymphoMotion follow the prescribed trajectories more faithfully and preserve scene stability across both settings. Their videos exhibit more accurate viewpoint transitions, stronger structural consistency, and fewer distortions during camera movement, demonstrating more reliable camera control, especially under object-level conditions.

\subsection{Independent Control of Object Motion.}
\label{sec:supplementary-object}

Additional examples in Fig.~\ref{fig:material-object} assess object-level control in static scenes. MotionCtrl exhibits inaccurate and unstable object behavior, failing to maintain coherent motion or consistent trajectories over time, even in the absence of camera movement. SymphoMotion, in contrast, adheres to the specified object-motion cues and generates smooth, consistent dynamics with noticeably improved temporal stability. These observations highlight the effectiveness of incorporating 3D trajectory conditioning, which enables more stable and reliable object-motion control.

\subsection{Simultaneous Control of Camera and Object Motions.}
\label{sec:supplementary-camera-object}

\begin{figure*}[t]
\centering
\includegraphics[width=1\textwidth]{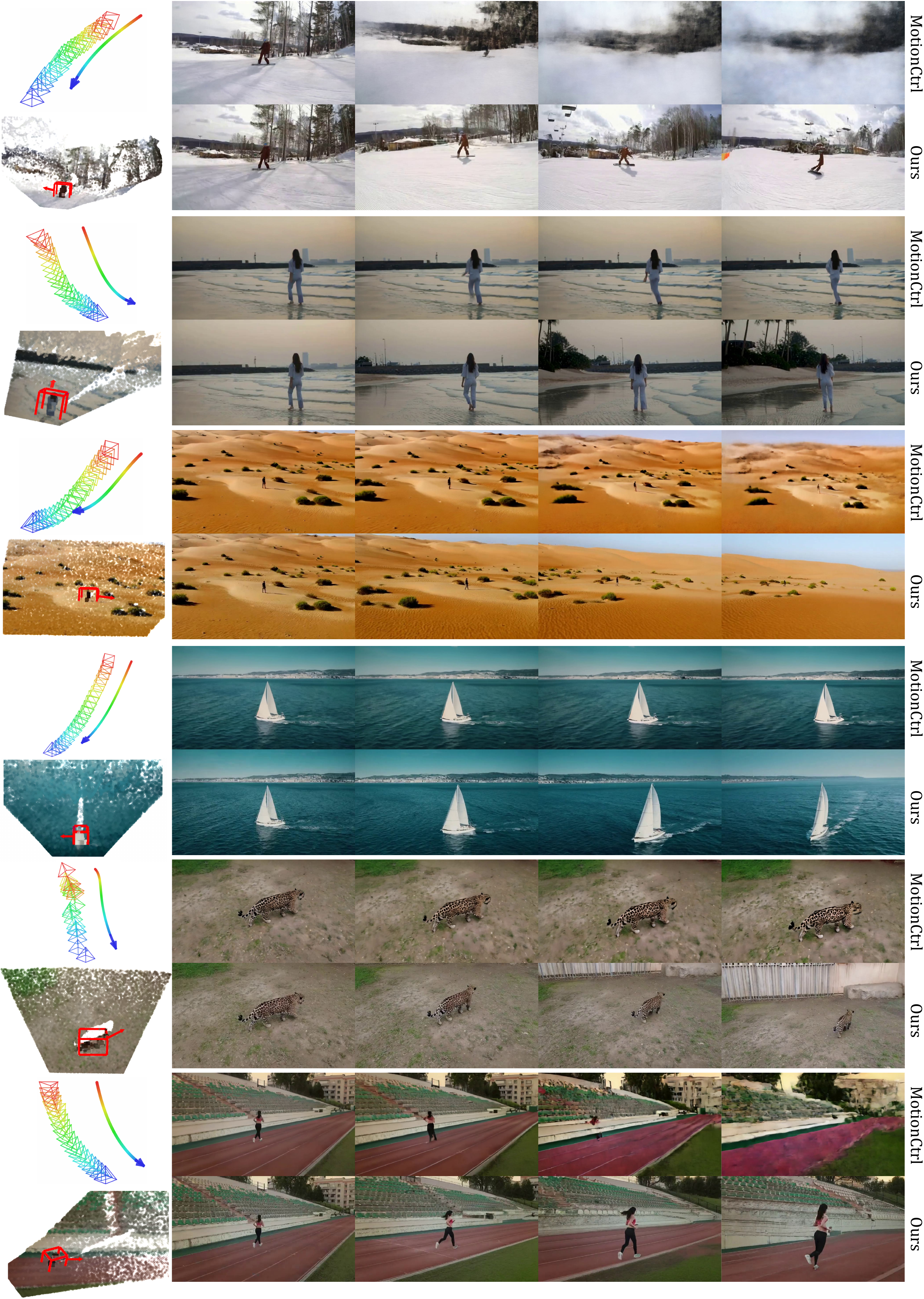}
\caption{
\textbf{More Qualitative Results on Simultaneous Camera and Object Motion Control.}
}
\label{fig:material-camera-object}
\end{figure*}

Fig.~\ref{fig:material-camera-object} examines the simultaneous control of camera and object motions. MotionCtrl exhibits pronounced limitations in this configuration, with both camera trajectories and object dynamics deviating from the intended motion cues. The method struggles to handle the interaction between the two motion sources, resulting in coupled and unstable behaviors. SymphoMotion, in contrast, coordinates both motions effectively, maintaining accurate camera movement while producing coherent object dynamics. These qualitative results highlight the robustness of our unified framework in handling coupled motion conditions.

\clearpage
{
    \small
    \bibliographystyle{ieeenat_fullname}
    \bibliography{main}
}

\end{document}